\documentclass[letterpaper, 10 pt, conference]{ieeeconf}  

\IEEEoverridecommandlockouts                              

\usepackage{cite}
\usepackage{amsmath,amssymb,amsfonts}
\usepackage{algorithmic}
\usepackage{graphicx}
\usepackage{tabularx} 
\usepackage{multirow}
\usepackage{textcomp}
\usepackage{xcolor}
\usepackage{diagbox}
\usepackage{subfigure}
\def\BibTeX{{\rm B\kern-.05em{\sc i\kern-.025em b}\kern-.08em
    T\kern-.1667em\lower.7ex\hbox{E}\kern-.125emX}}
\usepackage[colorlinks = true,
            linkcolor = black,
            urlcolor  = black,
            citecolor = black,
            anchorcolor = black]{hyperref}

\newcommand{\MYhref}[3][magenta]{\href{#2}{\color{#1}{#3}}}%

\title{\LARGE \bf
Fine-grained Hand Gesture Recognition in \\Multi-viewpoint Hand Hygiene
}

\author{Huy Q.Vo$^{1}$, Tuong Do$^{2}$, Vi C.Pham$^{1}$, Duy Nguyen$^{3}$, An T.Duong$^{1}$, and Quang D.Tran$^{1,2}$
\thanks{$^1$Willogy, Vietnam {\tt\small \{huy.vquoc, vi.pcao, an, quang\}@willogy.io}}
\thanks{$^2$AIOZ, Singapore {\tt\small \{tuong.khanh-long.do, quang.tran\}@aioz.io}}
\thanks{$^3$Blood Transfusion Hematology Hospital, Vietnam {\tt\small duynvm@bthh.org.vn}}
}

\begin{document}

\maketitle
\thispagestyle{empty}
\pagestyle{empty}

\begin{abstract}
This paper contributes a new high-quality dataset for hand gesture recognition in hand hygiene systems, named ``MFH". Generally, current datasets are not focused on: (i) fine-grained actions; and (ii) data mismatch between different viewpoints, which are available under realistic settings. To address the aforementioned issues, the MFH dataset is proposed to contain a total of 731147 samples obtained by different camera views in 6 non-overlapping locations.  
Additionally, each sample belongs to one of seven steps introduced by the World Health Organization (WHO). 
As a minor contribution,  inspired by advances in fine-grained image recognition and distribution adaptation,  this paper recommends using the self-supervised learning method to handle these preceding problems.
The extensive experiments on the benchmarking MFH dataset show that the introduced method yields competitive performance in both the Accuracy and the Macro F1-score. The code and the MFH dataset are available at \textit{\MYhref{https://github.com/willogy-team/hand-gesture-recognition-smc2021}{https://github.com/willogy-team/hand-gesture-recognition-smc2021}}.
\end{abstract}

\section{Introduction}
\label{sec:intro}

Hand hygiene, the so-called hand wash process, is an essential part that prevents infectious diseases in surgery. Even healthcare professionals frequently fail to follow the hand hygiene guidelines, hence, raise the chances of infection transmission. Fortunately, the hand hygiene technique is published by The World Health Organization (WHO) to help medical staffs keep all the surfaces of their hands clean during working. However, people may find it difficult to remember these steps correctly. Hence, it is substantial to automatically control the quality of the hand wash process in clinical environments. A computer vision system is one of the most efficient methods to accomplish this. Specifically, hand gesture recognition technologies, in particular, have already been used to assess hand hygiene compliance \cite{2ameling2011vision,6llorca2011vision,16ivanovs2020automated}.

In this paper, we consider the task of hand gesture recognition over a hand hygiene system. Our goal is to interpret the gestures of medical staff when they are washing their hands. 
However, different from other domains, hand gesture recognition for the hand wash process has two main aspects. First, hand hygiene is a process that contains fine-grained actions. It means that a deep learning agent is about to deal with both significant intra-class differences and subtle inter-class differences during their prediction process. For huge intra-class differences, the agent may recognize hand gestures that belong to the same category but may present significantly different poses and viewpoints\footnote{A viewpoint (or a scene) is the apparent distance and angle from which the camera views and records the subject}. For subtle inter-class differences, the agent may cope with gestures that belong to disparate steps but might be very similar apart from minor differences. Hand gesture recognition for fine-grained actions has been introduced in many approaches \cite{48lai2018cnn+,wang2016interacting,zhang2018handsense,dian2020towards}. However, in the hand hygiene domain, this aspect has not been focused on yet.

Second, the hand hygiene data are mismatched in distribution between the training phase and the inference phase. Indeed, the behavior of medical staff in different locations and camera views are not the same. Some steps are missing or not taken correctly, which requires lots of human effort to annotate. As a result, although similar data from other data distributions might be readily available, only a limited amount of data from the target distribution can be collected. Many works on hand gesture recognition have mentioned this data-driven issue \cite{45wu2020prototype,54jain2019synthetic,73rahimian2020fs}. However, there is no current dataset for the hand hygiene domain challenging enough to compare various gesture recognition methods.

By considering the two aspects above, our paper makes two contributions. The main contribution is to simulate both preceding problems by introducing a multi-viewpoint fine-grained hand hygiene dataset, named the MFH dataset (Fig.\ref{fig:7_steps_hand_washing}). 
It contains $731147$ samples in total, which are collected by $6$
camera views in $6$ different locations. All samples are split into $7$ classes in total. 
MFH dataset is distinguished from existing datasets in three aspects: the large intra-class difference, the subtle inter-class difference, and the data mismatch in distribution between the training phase and the inference phase. This dataset thus provides a more realistic benchmark. For performance evaluation, besides the accuracy, we recommend using the Macro F1-score for a more comprehensive measurement.

As a minor contribution, we address the preceding problems by applying the self-supervised learning approach for recognizing hand gestures in a hand hygiene system. Intuitively, the method is designed to maximize the mutual information between features extracted from multiple views of a shared context. This method is expected to deal with hand hygiene fine-grained problems and reduce the negative effect of distribution mismatch. To our knowledge, there is no previous approach that leverages self-supervised learning in dealing with multi-viewpoint hand gesture recognition.

Next, we review the related work in Section~\ref{Sec:Literature}. We then describe our dataset in Section~\ref{Sec:dataset} and the self-supervised learning method in Section~\ref{sec:self-supervised} . In Section~\ref{sec:exp}, we present extensive experimental results. Finally, we conclude the paper in Section~\ref{Sec:Conclusion}.

\section{Literature Review}
\label{Sec:Literature}

\textbf{Hand Gesture Recognition.}
Hand gesture recognition in a hand hygiene system is not a trivial learning task. There are two groups of approaches to coping with this. The former one, the sensor-based work, leverages the information from different sensors to establish the recognition \cite{7zhong2016washindepth,11fagert2017monitoring,13wang2020accurate,14mondol2020hawad}. The latter one, the image-based approach, mainly leverages input images and their correlated data to give out hand gesture predictions \cite{6llorca2011vision,16ivanovs2020automated}.

Specifically, in \cite{5singh2020automatic,7zhong2016washindepth,29cheng2015survey,30molchanov2015hand,48lai2018cnn+,61du2017hand, 27suarez2012hand}, the authors use information from depth sensors as input to give out hand gesture predictions. In \cite{23chen2019construct,25yang2019make,45wu2020prototype,46lai2020ensemble,48lai2018cnn+,59nguyen2019neural,62chen2017motion,69xie2019hgr}, the skeleton information is used as input.  Recently, Leap Motion has also been considered as an essential sensor for hand gesture recognition \cite{61du2017hand,68lupinetti20203d}. Although sensors play a crucial role in many situations, most sensors are costly and may not be easy to configure. Different from sensor-based approaches, the image-based ones mainly take images as the input. Particularly, in \cite{6llorca2011vision}, the authors extract HOG and HOF over images and then apply SVM for hand gesture classification.
In the most recent approach \cite{15dietz2018hand}, the authors propose hand hygiene monitoring based on the segmentation for separating hand parts of interacting and self-occluded hands.
To our knowledge, almost no works consider the fine-grained characteristic of hand gestures and the data mismatch over different viewpoints in the hand hygiene system during recognition. Moreover, datasets used in these previous works are private \cite{6llorca2011vision, 15dietz2018hand}, do not handle the fine-grained problem or the data mismatch problem \cite{16ivanovs2020automated}.

\textbf{Self-supervised Learning.}
Self-supervised learning (SSL) aims to self-generate robust representations from the unlabeled data according to the structure or characteristics of the data itself. SSL works as a supervision and benefits almost all types of downstream tasks, e.g., classification, recognition, or image retrieval \cite{kolesnikov2019revisiting,noroozi2018boosting,goyal2019scaling,chen2020big,si2020adversarial,lin2020ms2l,li2018self,han2020self,mundhenk2018improvements,wallace2020extending}. To deal effectively with the image classification task, the authors of \cite{zhai2019s4l} implement a hybrid system of self-supervised learning and semi-supervised learning. In \cite{misra2020self}, Pretext-Invariant Representation Learning (PIRL) is used to solve jigsaw puzzles and their rotation by enhancing the quality of the learned image representations. Recently, SSL is introduced to be more generalized since it can maximize the mutual information between features extracted from multiple views of a shared context \cite{bachman2019AmDim}. Inspired by the SSL, we leverage the generalization of the AmDim setup \cite{bachman2019AmDim} to deal with the data mismatch problem in hand gestures recognization from different viewpoints.

\section{The MFH dataset}
\label{Sec:dataset}
\subsection{Dataset description}
\begin{figure}[!t]
   \centering
\resizebox{\linewidth}{!}{
\setlength{\tabcolsep}{7pt}
\begin{tabular}{cccccccc}

\rotatebox[origin=l]{90}{\huge \textbf{Scene 1}} &
\shortstack{\huge \textbf{Step 1} \\ \includegraphics[width=0.3\linewidth]{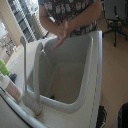}}&
\shortstack{\huge \textbf{Step 2} \\ \includegraphics[width=0.3\linewidth]{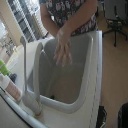}}&
\shortstack{\huge \textbf{Step 3} \\ \includegraphics[width=0.3\linewidth]{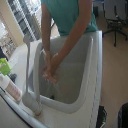}}&
\shortstack{\huge \textbf{Step 4} \\ \includegraphics[width=0.3\linewidth]{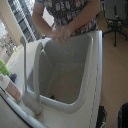}}&
\shortstack{\huge \textbf{Step 5} \\ \includegraphics[width=0.3\linewidth]{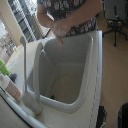}}&
\shortstack{\huge \textbf{Step 6} \\ \includegraphics[width=0.3\linewidth]{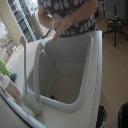}}&
\shortstack{\huge \textbf{Step 7} \\ \includegraphics[width=0.3\linewidth]{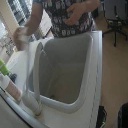}} \\ [7pt]
\hline\\
\rotatebox[origin=l]{90}{\huge \textbf{Scene 2}} &
\shortstack{\includegraphics[width=0.3\linewidth]{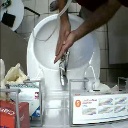}}&
\shortstack{\includegraphics[width=0.3\linewidth]{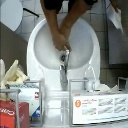}}&
\shortstack{\includegraphics[width=0.3\linewidth]{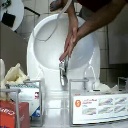}}&
\shortstack{\includegraphics[width=0.3\linewidth]{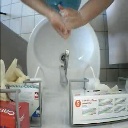}}&
\shortstack{\includegraphics[width=0.3\linewidth]{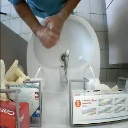}}&
\shortstack{\includegraphics[width=0.3\linewidth]{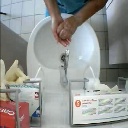}}&
\shortstack{\includegraphics[width=0.3\linewidth]{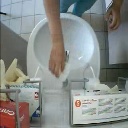}} \\ [7pt]
\hline\\
\rotatebox[origin=l]{90}{\huge \textbf{Scene 3}} &
\shortstack{\includegraphics[width=0.3\linewidth]{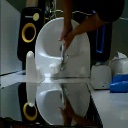}}&
\shortstack{\includegraphics[width=0.3\linewidth]{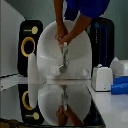}}&
\shortstack{\includegraphics[width=0.3\linewidth]{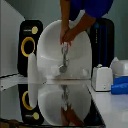}}&
\shortstack{\includegraphics[width=0.3\linewidth]{images/6Scenes_7Steps/Scene3/Step4_DataSet13_1_2020-06-27_12-06-39_camera105.avi-fix_frame1097.jpg}}&
\shortstack{\includegraphics[width=0.3\linewidth]{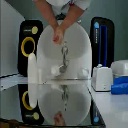}}&
\shortstack{\includegraphics[width=0.3\linewidth]{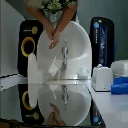}}&
\shortstack{\includegraphics[width=0.3\linewidth]{images/6Scenes_7Steps/Scene3/Step7_DataSet13_1_2020-06-27_13-50-18_camera105.avi-fix_frame934.jpg}} \\ [7pt]
\hline\\
\rotatebox[origin=l]{90}{\huge \textbf{Scene 4}} &
\shortstack{\includegraphics[width=0.3\linewidth]{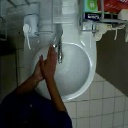}}&
\shortstack{\includegraphics[width=0.3\linewidth]{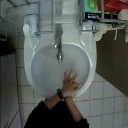}}&
\shortstack{\includegraphics[width=0.3\linewidth]{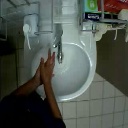}}&
\shortstack{\includegraphics[width=0.3\linewidth]{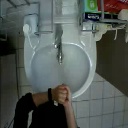}}&
\shortstack{\includegraphics[width=0.3\linewidth]{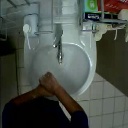}}&
\shortstack{\includegraphics[width=0.3\linewidth]{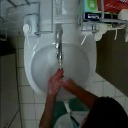}}&
\shortstack{\includegraphics[width=0.3\linewidth]{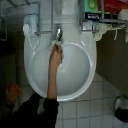}} \\ [7pt]
\hline\\
\rotatebox[origin=l]{90}{\huge \textbf{Scene 5}} &
\shortstack{\includegraphics[width=0.3\linewidth]{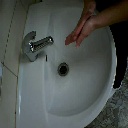}}&
\shortstack{\includegraphics[width=0.3\linewidth]{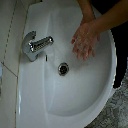}}&
\shortstack{\includegraphics[width=0.3\linewidth]{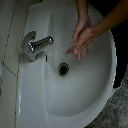}}&
\shortstack{\includegraphics[width=0.3\linewidth]{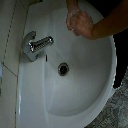}}&
\shortstack{\includegraphics[width=0.3\linewidth]{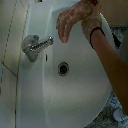}}&
\shortstack{\includegraphics[width=0.3\linewidth]{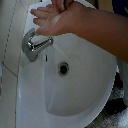}}&
\shortstack{\includegraphics[width=0.3\linewidth]{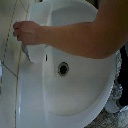}} \\ [7pt]
\hline\\
\rotatebox[origin=l]{90}{\huge \textbf{Scene 6}} &
\shortstack{\includegraphics[width=0.3\linewidth]{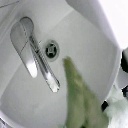}}&
\shortstack{\includegraphics[width=0.3\linewidth]{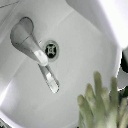}}&
\shortstack{\includegraphics[width=0.3\linewidth]{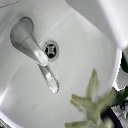}}&
\shortstack{\includegraphics[width=0.3\linewidth]{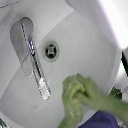}}&
\shortstack{\includegraphics[width=0.3\linewidth]{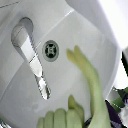}}&
\shortstack{\includegraphics[width=0.3\linewidth]{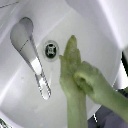}}&
\shortstack{\includegraphics[width=0.3\linewidth]{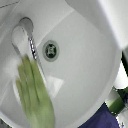}} \\ [7pt]

\end{tabular}
}
    \caption{Data samples of $6$ different viewpoints over $7$ hand hygiene steps in the MFH dataset.}
    \label{fig:7_steps_hand_washing} 
\end{figure}

The introduced MFH dataset is an on-top dataset using images from \cite{16ivanovs2020automated}. These images were collected monthly, and the total deployment duration was $3$ months.  During dataset collection, a total of $6$ cameras were placed in $6$ different locations.  All cameras are set at $640 \times 480$ pixels resolution, and their frame rate is 30 fps.
There are defined seven different hand washing movements as recommended by the WHO. These movements are as follows: palm to palm (Step $1$), palm over dorsum with fingers interlaced (Step $2$), palm to palm with fingers interlaced (Step $3$), back of fingers to opposing palm (Step $4$), rotational rubbing of the thumb (Step $5$), fingertips to palm (Step $6$), turning off the faucet with a paper towel (Step $7$). For more details, please visit Fig.\ref{fig:7_steps_hand_washing} which illustrates the visualization of these movements from different viewpoints. Additionally, it was necessary to identify whether a person is washing hands with a watch, a ring, or has lacquered nails. The reason is that these factors interfere with basic handwashing procedures and can be regarded as inappropriate for medical professionals in their work environment.

The dataset consists of  $1827$ annotated video files, each of which corresponds to a single hand-wash episode. The video files are split into frames that are easier to access. For each video file, there is a matching .json file, which contains the annotations of each frame in JSON format. Overlapping exists among all viewpoints (scenes); the dataset contains up to $731147$ samples (frames). 

Table \ref{tab:number_of_samples_MFH_dataset} illustrates the detail statistics of the MFH dataset. Specifically, each row denotes the number of samples of each specific scene. There are $6$ scenes in total. The column indicates $7$  steps in the hand wash process, and the row demonstrates the index of different scenes. 
Besides, we provide the total samples of each scene in the final column.
Through statistical, there is bias in the number of samples over classes in each viewpoint, i.e., imbalanced data. Moreover, the distribution of the bias over classes between these viewpoints is highly distinct. Our approach is expected to provide an opportunity to compare and evaluate the performance of different hand gesture recognition networks under two challenging aspects: fine-grained hand gestures and data distribution mismatch over different viewpoints. 
These aspects are consistent with practical usage. Hence, our dataset provides a testbed for methods applied in open systems.

\begin{table}[!ht]
\begin{center}
\setlength{\tabcolsep}{0.2em} 
{\renewcommand{\arraystretch}{1.5}
\resizebox{\linewidth}{!}{
\begin{tabular}{|c|c|c|c|c|c|c|c|c|}
\hline
    \diagbox[width=\dimexpr \textwidth/8+2\tabcolsep\relax, height=1cm]{ \textbf{Scene} }{ \textbf{Class} } & \textbf{Step 1} & \textbf{Step 2} & \textbf{Step 3} & \textbf{Step 4} & \textbf{Step 5} & \textbf{Step 6} & \textbf{Step 7} & \begin{tabular}[c]{@{}c@{}}\textbf{Total samples}\\\textbf{(per scene)}\end{tabular}\\ 
\hline
\textbf{1} & $10452$ & $6012$ & $2779$ & $2350$ & $1851$ & $962$ & $2084$ & $26490$ \\\hline
\textbf{2} & $77119$ & $74965$ & $63141$ & $57201$ & $72463$ & $53164$ & $21369$ & $419422$ \\\hline
\textbf{3} & $27148$ & $31221$ & $19275$ & $4068$ & $13372$ & $8333$ & $19569$ & $122986$ \\\hline
\textbf{4} & $13997$ & $15765$ & $8333$ & $11897$ & $8797$ & $9967$ & $14672$ & $83428$ \\\hline
\textbf{5} & $9978$ & $7186$ & $5739$ & $2997$ & $3306$ & $3026$ & $3273$ & $35505$ \\\hline
\textbf{6} & $11512$ & $9188$ & $6293$ & $1771$ & $2272$ & $2764$ & $9516$ & $43316$ \\
\hline
\end{tabular}
}
}
\end{center}
\caption{\label{tab:number_of_samples_MFH_dataset} The number of samples in each class of $6$ locations. 
}
\end{table}

\subsection{Evaluation protocols}
We typically use Accuracy to evaluate the effectiveness of different models in the hand gesture recognition task. Accuracy is calculated as the ratio between the number of correct predictions to the total number of predictions. The definition of Accuracy is also described as in (\ref{eq:acc}).

\begin{equation}
    \text{Accuracy} = \frac{C}{A}
    \label{eq:acc}
\end{equation}
where $C$ is the number of samples that are recognized correctly, $A$ is the number of all samples in the test data.

The Accuracy metric is essential in most cases. However, if the benchmarking dataset is not balanced, this metric has not much reference value. Since the MFH dataset is an imbalanced one, we introduce the Macro F1-score. Unlike Accuracy, which focuses on the importance of samples, the Macro F1-score puts the same importance on each class. The model that only performs well on the common classes while performing poorly on the rare classes will cause a low Macro F1-score.

Macro F1-score, so called Macro-averaged F1 score, is defined as the mean of class-wise/label-wise F1-scores. The F1-score ${F1}_i$ is the harmonic mean of precision and recall. Let $\text{TP}_i$, $\text{FP}_i$, $\text{FN}_i$, $\text{P}_i$, $\text{R}_i$ and $\text{F1}_i$ be the true positives, false positives, false negatives, precision, recall, F1-score with regard to class $i$ and $H$ is the harmonic mean. The precision $\text{P}_i$, the recall $\text{R}_i$, and the F1-score ${F1}_i$ are computed as in (\ref{eq:precision}), (\ref{eq:recall}), and (\ref{eq:f1-score}), respectively.

\begin{equation}
    \text{P}_i = \frac{\text{TP}_i}{\text{TP}_i + \text{FP}_i}
    \label{eq:precision}
\end{equation}
\begin{equation}
    \text{R}_i = \frac{\text{TP}_i}{\text{TP}_i + \text{FN}_i}
    \label{eq:recall}
\end{equation}
\begin{equation}
    \text{F1}_i = H(\text{P}_i, \text{R}_i) = 2*\frac{\text{P}_i*\text{R}_i}{\text{P}_i + \text{R}_i}
    \label{eq:f1-score}
\end{equation}
where $N$ is the number of classes, $\text{F1}_i$ is the calculated F1 value on class $i$ $(1 \leq i \leq N)$. 

In the final step, the Macro F1-score is then calculated using (\ref{eq:macro-f1}).
\begin{equation}
    \text{Macro F1-score} = \frac{1}{N} \sum_{i=1}^{N} {\text{F1-score}_i}
    \label{eq:macro-f1}
\end{equation}

\subsection{Comparing with existing datasets}
A statistics comparison with existing datasets is shown in Table \ref{tab:datasets}. Our dataset contains $731147$ samples, which is larger than the current largest dataset \cite{16ivanovs2020automated} by double. Different from \cite{15dietz2018hand} which uses Depth Infrared images as inputs, RGB images are leveraged in our introduced MFH dataset since they provide good visual information. The highlight of the MFH dataset, also the key difference when comparing MFH with other datasets, is that it contains $6$ sub-datasets from $6$ non-overlapping viewpoints. Under realistic Healthcare Industry settings, our dataset serves as an ideal benchmark for learning methods that focus on the generalization capacities and data mismatching by opening two different evaluation protocols for testing.

\begin{table}[!ht]
\centering
\setlength{\tabcolsep}{0.2 em} 
\resizebox{\linewidth}{!}{
{\renewcommand{\arraystretch}{1.2}
\begin{tabular}{|l|c|c|c|c|}
\hline
\multicolumn{1}{|l|}{\textbf{Datasets}} & \textbf{MFH}                                                                  & \textbf{HWQA \cite{6llorca2011vision}} & \textbf{SIH\cite{15dietz2018hand}} & \textbf{AQA\cite{16ivanovs2020automated}} \\ \hline
Type of Input                  & RGB image                                                            & RGB image                                      & Depth Infrared image                                & RGB image                                         \\ \hline
Num. Frames                    & $731147$                                                               & $8408$                                          & $83000$                                      & $309315$                                            \\ \hline
Num. Gestures                  & $7$                                                                    & $7^*$                                              & $7$                                          & $8^*$                                                \\ \hline
Num. View Split         & $6$                                                                  & $1$                                             & $1$                                         & $1$                                                \\ \hline
Evaluation Protocol            & \begin{tabular}[c]{@{}c@{}}Accuracy \&\\ Macro F1-score\end{tabular} & Accuracy                                       & Accuracy                                   & Accuracy                                          \\ \hline
\end{tabular}
}
}
\caption{Comparing MFH with existing datasets \cite{6llorca2011vision,15dietz2018hand,16ivanovs2020automated}.$*$ indicates the dataset contained the ``other" image label that do not belong to ones introduced by WHO.
}
\label{tab:datasets}
\end{table}
\section{Self-supervised learning}
\label{sec:self-supervised}
Self-supervised learning derives from unsupervised learning and can be applied in any recognition or classification task. It aims to learn semantically meaningful representations from unlabeled data. Generally, some portion of the data is retained, and the network is tasked with predicting it. One of the most effective approaches is to design a pretext task, which maximizes the mutual information between features extracted from multiple views of a shared context. The context here is the input images, and the preceding views are augmented from these inputs.

By following \cite{bachman2019AmDim, chen2021self}, we determine the mutual information (MI), which measures the shared information between two random variables $X$ and $Y$. MI is defined as the Kullback-Leiber (KL) divergence between the joint $P(x,y)$ and the product of the marginals $P(x)$ and $P(y)$.
\begin{equation}
\begin{split}
I(X,Y) & = D_{KL}(p(x,y)||p(x)p(y)) \\
& = \sum \sum p(x,y)log(\frac{p(x|y)}{p(x)})
\end{split}
\label{eq:IXY}
\end{equation}
Since it is not easy to direct access to the underlying distribution to estimate MI, we instead maximize a lower bound on MI by minimizing the Noise Contrastive Estimate (NCE) loss based on negative sampling.

Our objective is to maximize MI between global features and local features from two views $(x_a,x_b)$ of the same hand gesture input image .i.e., $\langle f_g(x_a),f_5(x_b) \rangle$, $\langle f_g(x_a),f_7(x_b) \rangle$ and $\langle f_5(x_a),f_5(x_b) \rangle$. Where $f_g$, $f_5$, $f_7$ are the global feature, the encoder's $5 \times 5$ local feature map and the encoder's $7 \times 7$ feature map respectively. The NCE loss between $f_i(x_a)$ and $f_j(x_b)$ is defined in  (\ref{eq:loss_sst_sigle}).

\begin{multline}
    \mathcal{L}_{ssl}(f_i(x_a),f_j(x_b))= \\
    -log \frac{exp\{\phi(f_i(x_a),f_j(x_b))\}}{\sum_{\widetilde{x_b}\in \mathcal{N}_x\cup x_b}exp\{\phi(f_i(x_a),f_j(\widetilde{x_b}))\}}
    \label{eq:loss_sst_sigle}    
\end{multline}
where $\mathcal{N}_x$ are the negative samples of image $x$, $\phi$ is the distance metric function. 

The overall loss between $x_a$ and $x_b$ is the total of the NCE losses and is written  in (\ref{eq:final_loss}).
\begin{equation}
\begin{split}
    \mathcal{L}_{ssl}(x_a,x_b)=\mathcal{L}_{ssl}(f_g(x_a),f_5(x_b))+\mathcal{L}_{ssl}(f_g(x_a),f_7(x_b)) \\
    +\mathcal{L}_{ssl}(f_5(x_a),f_5(x_b))
    \label{eq:final_loss}
\end{split}
\end{equation}

It is worth noting that, after finishing the training process, the self-supervised learning network is leveraged as an encoder to extract features for the further classification task. To achieve hand gesture recognition, we need to train a supervised learning network, i.e., a classifier, on top of the aforementioned extracted features using the annotated hand gesture data. The structure of the self-supervised learning network is the standard ResNet \cite{he2016deep}. For the classifier, a linear layer or a multilayer perceptron is used as the structure. For more details about these structures, please visit \cite{bachman2019AmDim}.

\section{Experiments}
\label{sec:exp}
\subsection{Implementation details, data setup and baselines}
\textbf{Implementation details.} All experiments are conducted on an NVIDIA Titan V GPU with $12$GB RAM. All models are trained by using Stochastic Gradient Descent with a momentum of $0.9$. The initial learning rate is set to $0.001$, with exponential decay of $0.9$ after every two epochs. The maximum number of epochs is set at $40$.

\textbf{Data setup.} 
In the MFH dataset, the data got from each viewpoint is split into a train set and a test set. Each set contains $50$ percent of data, and samples in each are not overlapping. There are two scenarios for the evaluation phase. The first scenario is that the model is trained and tested within the same camera data, i.e., ``same scenes" scenario. The second scenario is that the model is trained in a specific scene notwithstanding its effectiveness is evaluated in the data collected from other scenes, i.e., ``cross scenes" scenario.

\textbf{Baselines.} 
MobileNetV2 \cite{sandler2018mobilenetv2}, ResNet-18 \cite{he2016deep}, and InceptionNetV3 \cite{szegedy2016rethinking} are leveraged as the baseline network for our analysis. These models are pretrained on the Imagenet dataset \cite{deng2009imagenet} and then fine-tuned in a specific sub-dataset so as to maximize their performance. 
AmDim \cite{bachman2019AmDim}, a self-supervised representation learning baseline, is expected to work well under the limitation of our dataset. Following the setup of other baselines, the model is pre-trained on the Imagenet dataset \cite{deng2009imagenet}. Finally, we train a classification on top of the features for the recognition of hand gestures.
\subsection{Experimental details}

\begin{figure}[!ht]
  \centering
  \includegraphics[width=0.5\linewidth]{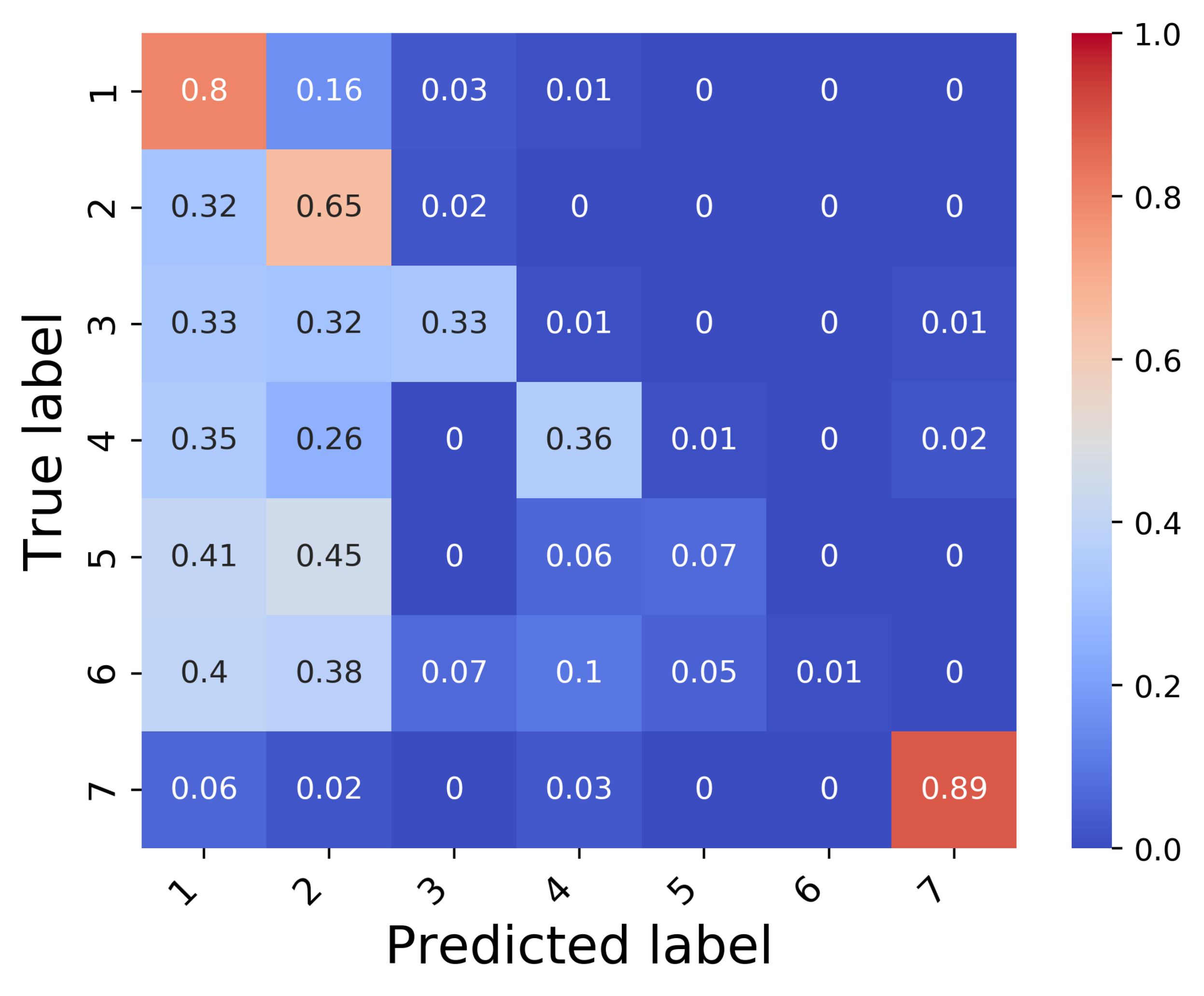}
  \caption{The confusion matrix of predicted samples in different classes over $1$-st scene data of MFH dataset.}
 \label{fig:step_analysis}
\end{figure}

\textbf{Fine-grained action analysis}
To identify the effectiveness of deep network over hand wash actions, we use the well-known InceptionV3, which is pre-trained in the ImageNet dataset\cite{deng2009imagenet}, as the baseline. An analysis is established in the $1$-st scene data of the MFH dataset. In Figure \ref{fig:step_analysis}, we present the confusion matrix over different tested samples. The results imply that there has a visible of confusing predictions over classes, e.g., The model tends to give out the predicted Step $1,2$ when it meets images from Step $5,6$. The main reason is that gestures that belong to disparate steps might be very similar apart from some minor differences.
\begin{figure}[!ht]
  \centering
    \subfigure[]{\includegraphics[width=0.49\linewidth]{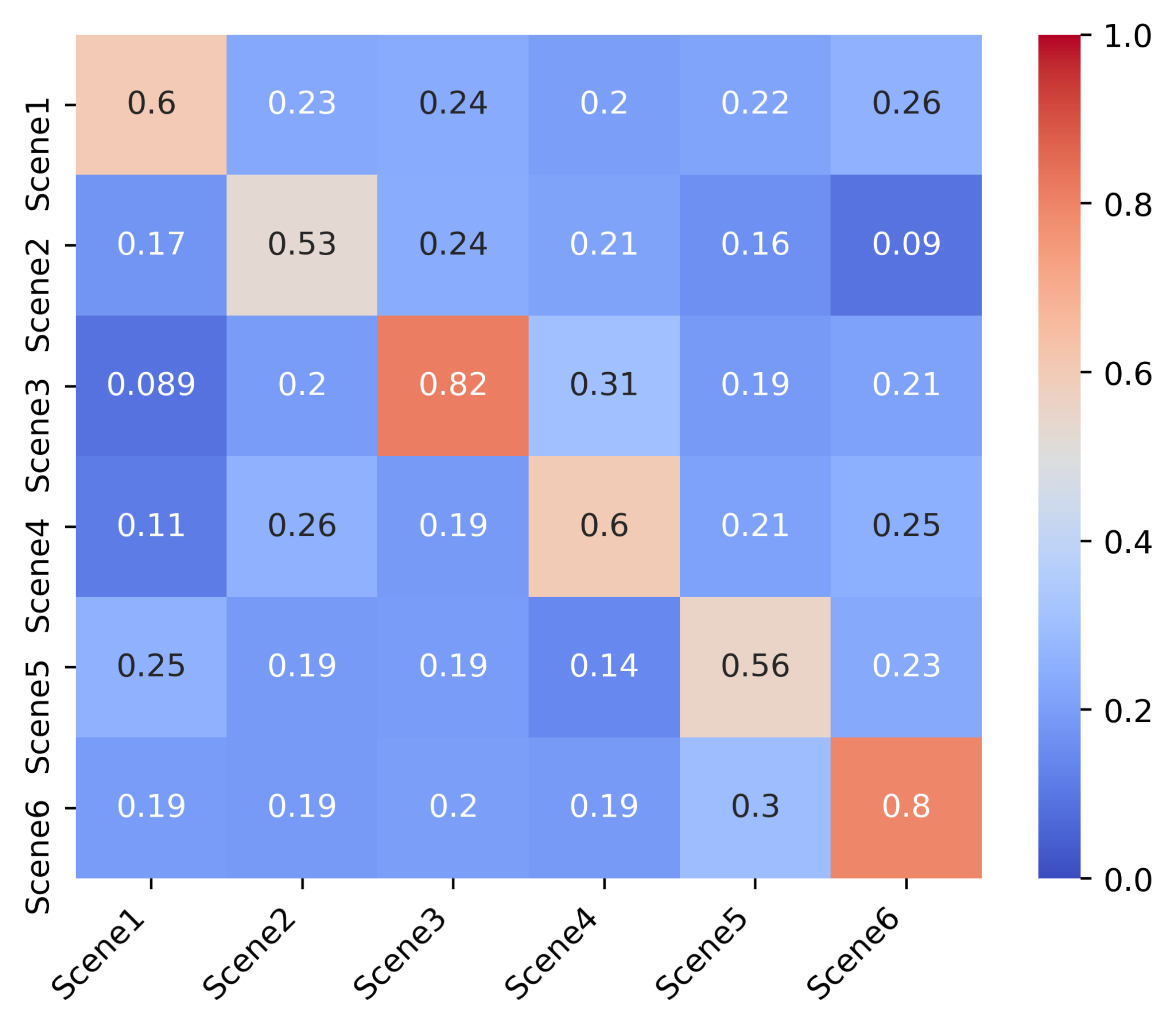}}
    \subfigure[]{\includegraphics[width=0.49\linewidth]{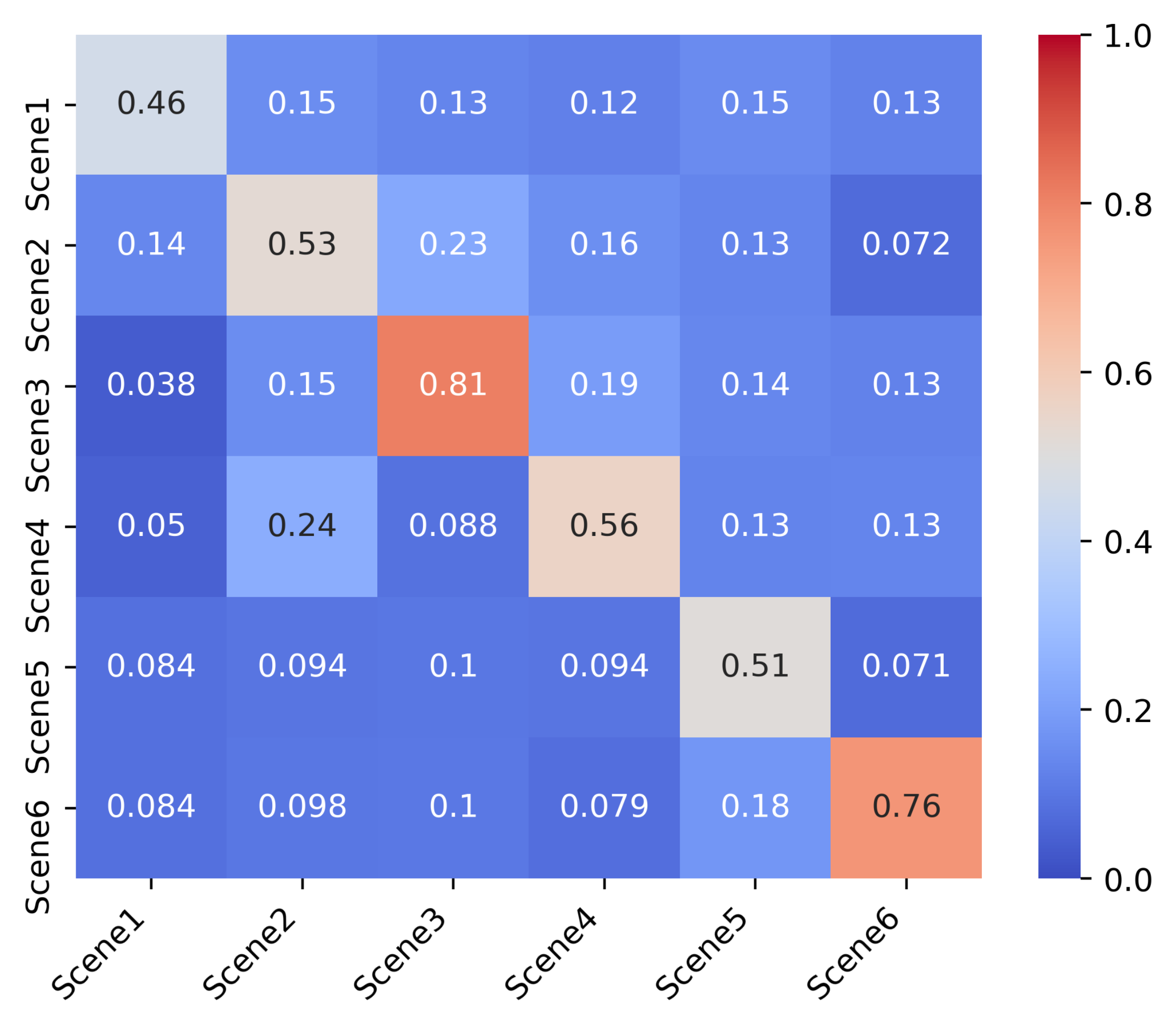}}
 \caption{The performance of InceptionV3 in $6$ viewpoints of the MFH dataset. \textbf{(a)} for Accuracy metric, and \textbf{(b)} for Macro F1-score.}
 \label{fig:InceptionV3_performance}
\end{figure}

\textbf{Data mismatch analysis.} To further understand the MFH dataset and its challenges, we provide the recognition results between all scene pairs in Figure \ref{fig:InceptionV3_performance} where InceptionV3 is used as the baseline. Each value in the figure is the Accuracy( Figure \ref{fig:InceptionV3_performance}- a) or  Macro F1-score (Figure \ref{fig:InceptionV3_performance} - b) when we test a specific model in the corresponding test set. Note that the row denotes the train data, and the column denotes the test data. Through the figure, hand gesture recognition within the same camera view, i.e., the ``same scenes" testing scenario, yields the highest accuracy score. On the other hand, as expected, the performance among different camera pairs,  i.e., the ``cross scenes" testing scenario, varies a lot. In most cases, InceptionV3 achieves low results due to the mismatch in distribution between train set and test set, regardless of the number of classes are not much, and the network itself is strong enough. Besides, the Macro F1-score is by far lower than the Accuracy in most ``cross scenes" experiment results.
It indicates that the imbalance over classes further increases the data mismatch between viewpoints.

Table \ref{tab:baseline} demonstrates the results obtained by different typical deep network structures including Mobilenet \cite{sandler2018mobilenetv2}, ResNet \cite{he2016deep}, and InceptionNet \cite{szegedy2016rethinking}. All of the preceding networks do not achieve good results during the inference phase of ``cross scenes" in both the Accuracy and the Macro F1-score. These results imply that all benchmarking networks can not work well with the data mismatch problem.

\begin{table}[!ht]
\centering
\setlength{\tabcolsep}{0.2 em} 
\resizebox{\linewidth}{!}{
{\renewcommand{\arraystretch}{1.5}
\begin{tabular}{|c|c|c|c|c|}
\hline
\multirow{2}{*}{\textbf{Method}} & \multicolumn{2}{c|}{\textbf{Avg. Accuracy}}  & \multicolumn{2}{c|}{\textbf{Avg. Macro F1}}  \\ \cline{2-5} 
                                 & \textit{same scenes} & \textit{cross scenes} & \textit{same scenes} & \textit{cross scenes} \\ \hline
MobilenetV2 \cite{sandler2018mobilenetv2}                      & $0.63 \pm 0.018$ & $0.18 \pm 0.003$ & $0.55 \pm 0.040$ & $0.09 \pm 0.001$ \\ \hline
ResNet-18 \cite{he2016deep}                       &      $0.58 \pm 0.003$      &     $0.22 \pm 0.003$    &     $0.47 \pm 0.018$      &    $0.10 \pm 0.001$               \\ \hline
InceptionV3 \cite{szegedy2016rethinking}                      & $0.65 \pm 0.013$                     & $0.20 \pm 0.003$                      & $0.61 \pm 0.018$                      & $0.12 \pm 0.002$                      \\ \hline
\textbf{AmDim \cite{bachman2019AmDim}}                            & $\mathbf{0.84 \pm 0.003}$ & $\mathbf{0.30 \pm 0.009}$ & $\mathbf{0.83 \pm 0.005}$ & $\mathbf{0.25 \pm 0.010}$ \\ \hline
\end{tabular}
}
}
\caption{The average performance comparison between different deep learning networks over $6$ viewpoints of MFH data.
}
\label{tab:baseline}
\end{table}

\begin{figure}[!h]
  \centering
    \subfigure[]{\includegraphics[width=0.49\linewidth]{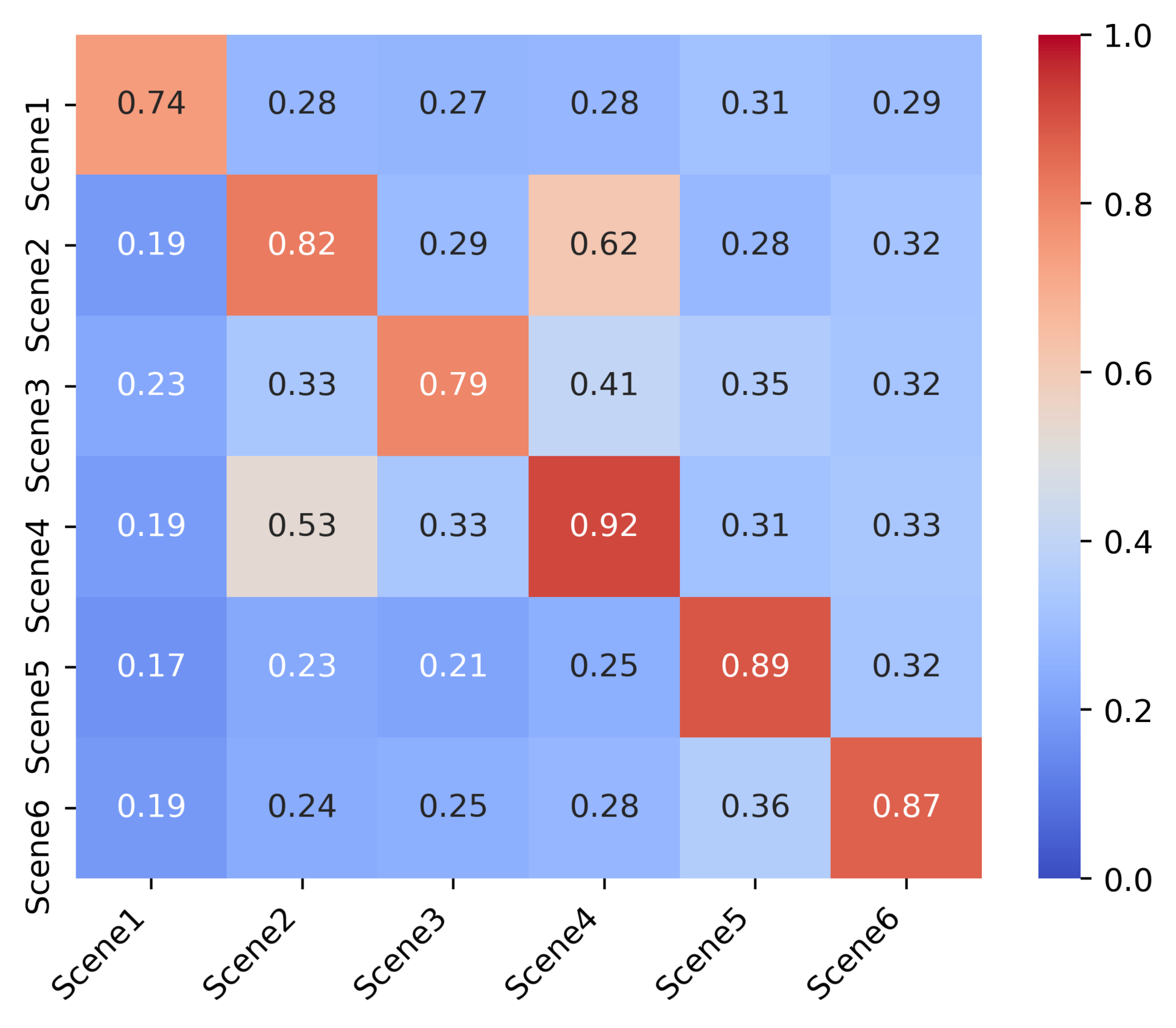}}
    \subfigure[]{\includegraphics[width=0.49\linewidth]{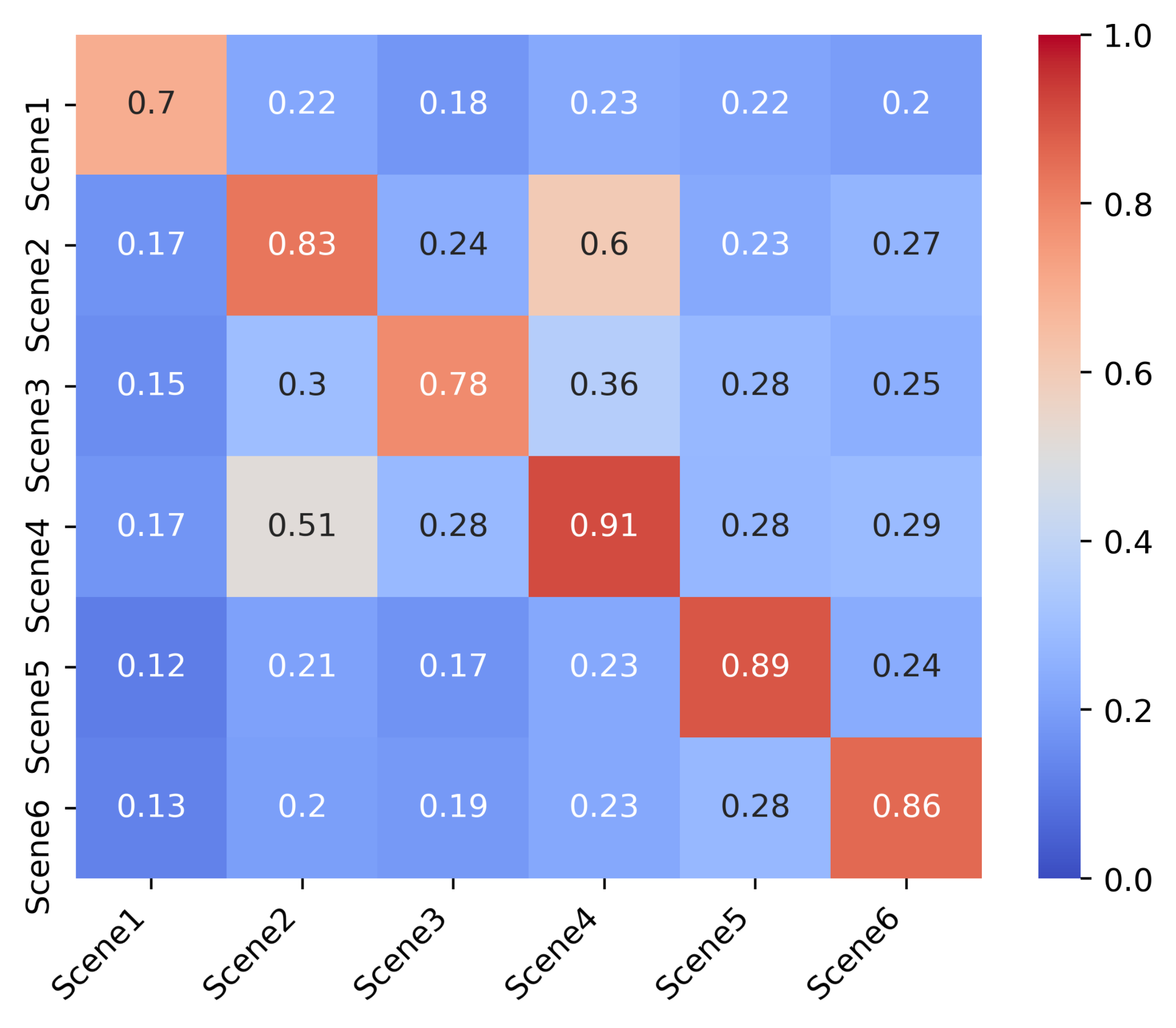}}
 \caption{The performance of AmDim in $6$ viewpoints of the MFH dataset. \textbf{(a)} illustrates the performance in Accuracy, and \textbf{(b)} demonstrates the performance in Macro F1-score. 
 }
 \label{fig:AmDim_performance}
\end{figure}

\textbf{Self-supervised learning analysis}
The AmDim \cite{bachman2019AmDim} is leveraged as a self-supervised learning baseline to deal with both analyzed problems. Table \ref{tab:baseline} demonstrates the performance comparison between AmDim and other learning methods. 
In the ``same scenes" testing scenario, AmDim outperforms other networks by a large margin. This result indicates that AmDim can learn robust features for recognizing fine-grained hand gestures.  
In the ``cross scenes" testing scenario, AmDim also achieves significant improvements in the Accuracy metric. Especially when comparing it with InceptionV3 - the most effective baseline, i.e., the Avg. The accuracy gap between AmDim and InceptionV3  is $0.1$. Hence, the self-supervised learning approach can deal with the mismatch in data distribution over different viewpoints.  It is worth noting that AmDim also outperforms other baselines in Macro F1-score, which validate the effectiveness of AmDim over imbalanced data problems (See Figure \ref{fig:AmDim_performance} for quantitative results of AmDim in over $6$ viewpoints.) 

Table \ref{tab:abl} presents the AmDim performance with and without pretraining the model in the ImageNet dataset. In terms of the ``cross-scene" scenario, the results show that both setups give better scores than the baseline InceptionV3 in two metrics, regardless that InceptionV3 is pre-trained on the ImageNet dataset. This again validates the effectiveness of self-supervised learning. Besides, through empirical experiments, we have investigated that using a deeper neural network, e.g., the multilayer perceptron (MLP), for classification achieves better results than using a linear one.
\begin{table}[!ht]
\centering
\setlength{\tabcolsep}{0.08 em} 
{\renewcommand{\arraystretch}{1.2}
\begin{tabular}{|c|c|c|c|c|c|c|}
\hline
\multirow{3}{*}{\textbf{Method}}                                 & \multirow{3}{*}{\textbf{\begin{tabular}[c]{@{}c@{}}Pretrained in\\ ImageNet\end{tabular}}} & \multirow{3}{*}{\textbf{\begin{tabular}[c]{@{}c@{}}Classifier\end{tabular}}} & \multicolumn{2}{c|}{\textbf{Avg. Accuracy}}                                                                                        & \multicolumn{2}{c|}{\textbf{Avg. Macro F1}}                                                                                        \\ \cline{4-7} 
                                                                 &                                                                                            &                                                                                           & \textit{\begin{tabular}[c]{@{}c@{}}same \\ scenes\end{tabular}} & \textit{\begin{tabular}[c]{@{}c@{}}cross \\ scenes\end{tabular}} & \textit{\begin{tabular}[c]{@{}c@{}}same \\ scenes\end{tabular}} & \textit{\begin{tabular}[c]{@{}c@{}}cross\\ scenes\end{tabular}} \\ \hline
\begin{tabular}[c]{@{}c@{}}InceptionV3 \cite{szegedy2016rethinking}\end{tabular} & Yes                                                                                        & \_                                                                                        & \multicolumn{1}{c|}{0.65}                                         & \multicolumn{1}{c|}{0.20}                                            & \multicolumn{1}{c|}{0.61}                                           & \multicolumn{1}{c|}{0.12}                                            \\ \hline
\multirow{4}{*}{AmDim \cite{bachman2019AmDim}}                                           & No                                                                                         & Linear                                                                                    &  $0.39$                                                               &   $0.21$                                                               &    $0.26$                                                             &    $0.13$                                                              \\ \cline{2-7} 
                                                                 & Yes                                                                                        & Linear                                                                                    & $0.73$                                                                & $0.29$                                                                 & $0.71$                                                               & $0.24$                                                                 \\ \cline{2-7} 
                                                                 & No                                                                                         & MLP                                                                                       &  $0.60$                                                               &    $0.22$                                                              &      $0.54$                                                           &    $0.15$                                                              \\ \cline{2-7} 
                                                                 & Yes                                                                                        & MLP                                                                                       & $0.84$                                                                & $0.30$                                                                 & $0.83$                                                                & $0.25$                                                                 \\ \hline
\end{tabular}
}
\caption{The average performance comparison between different AmDim setup over $6$ viewpoints of MFH data.
}
\label{tab:abl}
\end{table}

\section{Conclusion}
\label{Sec:Conclusion}
We introduce a multi-viewpoint fine-grained hand hygiene dataset (MFH) that reaches closer to realistic settings,  especially in the Healthcare Industry.  Our new dataset will enable research possibilities in multiple directions, e.g., deep learning, fine-grained learning, multi-view learning, and data distribution learning. Besides, self-supervised learning (SSL) is presented to deal with fine-grained hand gestures and data mismatch problems.  The extensive experiments show that SSL yields the best performance with various competitive baselines in Accuracy and Macro F1-score.

\section*{ACKNOWLEDGMENT}
Special thanks to AIOZ Singapore company and Blood Transfusion Hematology Hospital Vietnam for the valuable support on the cooperation.


\bibliographystyle{ieeetr}
\bibliography{refs}

\begin{thebibliography}{10}

\bibitem{2ameling2011vision}
S.~Ameling, J.~Li, J.~Zhou, A.~Ghosh, G.~Lacey, E.~Creamer, and H.~Humphreys,
  ``A vision-based system for handwashing quality assessment with realtime
  feedback,'' in {\em The Eighth IASTED International Conference on Biomedical
  Engineering, Biomed 2011, Innsbruck, Austria}, 2011.

\bibitem{6llorca2011vision}
D.~F. Llorca, I.~Parra, M.~{\'A}. Sotelo, and G.~Lacey, ``A vision-based system
  for automatic hand washing quality assessment,'' {\em Machine Vision and
  Applications}, vol.~22, no.~2, pp.~219--234, 2011.

\bibitem{16ivanovs2020automated}
M.~Ivanovs, R.~Kadikis, M.~Lulla, A.~Rutkovskis, and A.~Elsts, ``Automated
  quality assessment of hand washing using deep learning,'' {\em arXiv preprint
  arXiv:2011.11383}, 2020.

\bibitem{48lai2018cnn+}
K.~Lai and S.~N. Yanushkevich, ``Cnn+ rnn depth and skeleton based dynamic hand
  gesture recognition,'' in {\em 2018 24th International Conference on Pattern
  Recognition (ICPR)}, pp.~3451--3456, IEEE, 2018.

\bibitem{wang2016interacting}
S.~Wang, J.~Song, J.~Lien, I.~Poupyrev, and O.~Hilliges, ``Interacting with
  soli: Exploring fine-grained dynamic gesture recognition in the
  radio-frequency spectrum,'' in {\em Proceedings of the 29th Annual Symposium
  on User Interface Software and Technology}, pp.~851--860, 2016.

\bibitem{zhang2018handsense}
Z.~Zhang, Z.~Tian, and M.~Zhou, ``Handsense: smart multimodal hand gesture
  recognition based on deep neural networks,'' {\em Journal of Ambient
  Intelligence and Humanized Computing}, pp.~1--16, 2018.

\bibitem{dian2020towards}
C.~Dian, D.~Wang, Q.~Zhang, R.~Zhao, and Y.~Yu, ``Towards domain-independent
  complex and fine-grained gesture recognition with rfid,'' {\em Proceedings of
  the ACM on Human-Computer Interaction}, vol.~4, no.~ISS, pp.~1--22, 2020.

\bibitem{45wu2020prototype}
J.~Wu, Y.~Zhang, and X.~Zhao, ``A prototype-based generalized zero-shot
  learning framework for hand gesture recognition,'' {\em arXiv preprint
  arXiv:2009.13957}, 2020.

\bibitem{54jain2019synthetic}
V.~Jain, S.~Aggarwal, S.~Mehta, and R.~Hebbalaguppe, ``Synthetic video
  generation for robust hand gesture recognition in augmented reality
  applications,'' {\em arXiv preprint arXiv:1911.01320}, 2019.

\bibitem{73rahimian2020fs}
E.~Rahimian, S.~Zabihi, A.~Asif, D.~Farina, S.~F. Atashzar, and A.~Mohammadi,
  ``Fs-hgr: Few-shot learning for hand gesture recognition via
  electromyography,'' {\em arXiv preprint arXiv:2011.06104}, 2020.

\bibitem{7zhong2016washindepth}
H.~Zhong, S.~S. Kanhere, and C.~T. Chou, ``Washindepth: Lightweight hand wash
  monitor using depth sensor,'' in {\em Proceedings of the 13th International
  Conference on Mobile and Ubiquitous Systems: Computing, Networking and
  Services}, pp.~28--37, 2016.

\bibitem{11fagert2017monitoring}
J.~Fagert, M.~Mirshekari, S.~Pan, P.~Zhang, and H.~Y. Noh, ``Monitoring
  hand-washing practices using structural vibrations,'' {\em Structural Health
  Monitoring}, 2017.

\bibitem{13wang2020accurate}
C.~Wang, Z.~Sarsenbayeva, X.~Chen, T.~Dingler, J.~Goncalves, and V.~Kostakos,
  ``Accurate measurement of handwash quality using sensor armbands: Instrument
  validation study,'' {\em JMIR mHealth and uHealth}, vol.~8, no.~3, p.~e17001,
  2020.

\bibitem{14mondol2020hawad}
M.~A.~S. Mondol and J.~A. Stankovic, ``Hawad: Hand washing detection using
  wrist wearable inertial sensors,'' in {\em 2020 16th International Conference
  on Distributed Computing in Sensor Systems (DCOSS)}, pp.~11--18, IEEE, 2020.

\bibitem{5singh2020automatic}
A.~Singh, A.~Haque, A.~Alahi, S.~Yeung, M.~Guo, J.~R. Glassman, W.~Beninati,
  T.~Platchek, L.~Fei-Fei, and A.~Milstein, ``Automatic detection of hand
  hygiene using computer vision technology,'' {\em Journal of the American
  Medical Informatics Association}, vol.~27, no.~8, pp.~1316--1320, 2020.

\bibitem{29cheng2015survey}
H.~Cheng, L.~Yang, and Z.~Liu, ``Survey on 3d hand gesture recognition,'' {\em
  IEEE transactions on circuits and systems for video technology}, vol.~26,
  no.~9, pp.~1659--1673, 2015.

\bibitem{30molchanov2015hand}
P.~Molchanov, S.~Gupta, K.~Kim, and J.~Kautz, ``Hand gesture recognition with
  3d convolutional neural networks,'' in {\em Proceedings of the IEEE
  conference on computer vision and pattern recognition workshops}, pp.~1--7,
  2015.

\bibitem{61du2017hand}
Y.~Du, S.~Liu, L.~Feng, M.~Chen, and J.~Wu, ``Hand gesture recognition with
  leap motion,'' {\em arXiv preprint arXiv:1711.04293}, 2017.

\bibitem{27suarez2012hand}
J.~Suarez and R.~R. Murphy, ``Hand gesture recognition with depth images: A
  review,'' in {\em 2012 IEEE RO-MAN: the 21st IEEE international symposium on
  robot and human interactive communication}, pp.~411--417, IEEE, 2012.

\bibitem{23chen2019construct}
Y.~Chen, L.~Zhao, X.~Peng, J.~Yuan, and D.~N. Metaxas, ``Construct dynamic
  graphs for hand gesture recognition via spatial-temporal attention,'' {\em
  arXiv preprint arXiv:1907.08871}, 2019.

\bibitem{25yang2019make}
F.~Yang, Y.~Wu, S.~Sakti, and S.~Nakamura, ``Make skeleton-based action
  recognition model smaller, faster and better,'' in {\em Proceedings of the
  ACM Multimedia Asia}, 2019.

\bibitem{46lai2020ensemble}
K.~Lai and S.~Yanushkevich, ``An ensemble of knowledge sharing models for
  dynamic hand gesture recognition,'' in {\em 2020 International Joint
  Conference on Neural Networks (IJCNN)}, pp.~1--7, IEEE, 2020.

\bibitem{59nguyen2019neural}
X.~S. Nguyen, L.~Brun, O.~L{\'e}zoray, and S.~Bougleux, ``A neural network
  based on spd manifold learning for skeleton-based hand gesture recognition,''
  in {\em Proceedings of the IEEE/CVF Conference on Computer Vision and Pattern
  Recognition}, pp.~12036--12045, 2019.

\bibitem{62chen2017motion}
X.~Chen, H.~Guo, G.~Wang, and L.~Zhang, ``Motion feature augmented recurrent
  neural network for skeleton-based dynamic hand gesture recognition,'' in {\em
  2017 IEEE International Conference on Image Processing (ICIP)},
  pp.~2881--2885, IEEE, 2017.

\bibitem{69xie2019hgr}
H.~Xie, J.~Wang, B.~Shao, J.~Gu, and M.~Li, ``Le-hgr: A lightweight and
  efficient rgb-based online gesture recognition network for embedded ar
  devices,'' in {\em 2019 IEEE International Symposium on Mixed and Augmented
  Reality Adjunct (ISMAR-Adjunct)}, pp.~274--279, IEEE, 2019.

\bibitem{68lupinetti20203d}
K.~Lupinetti, A.~Ranieri, F.~Giannini, and M.~Monti, ``3d dynamic hand gestures
  recognition using the leap motion sensor and convolutional neural networks,''
  in {\em International Conference on Augmented Reality, Virtual Reality and
  Computer Graphics}, pp.~420--439, Springer, 2020.

\bibitem{15dietz2018hand}
A.~Dietz, A.~P{\"o}sch, and E.~Reithmeier, ``Hand hygiene monitoring based on
  segmentation of interacting hands with convolutional networks,'' in {\em
  Medical Imaging 2018: Imaging Informatics for Healthcare, Research, and
  Applications}, vol.~10579, p.~1057914, International Society for Optics and
  Photonics, 2018.

\bibitem{kolesnikov2019revisiting}
A.~Kolesnikov, X.~Zhai, and L.~Beyer, ``Revisiting self-supervised visual
  representation learning,'' in {\em Proceedings of the IEEE/CVF Conference on
  Computer Vision and Pattern Recognition}, pp.~1920--1929, 2019.

\bibitem{noroozi2018boosting}
M.~Noroozi, A.~Vinjimoor, P.~Favaro, and H.~Pirsiavash, ``Boosting
  self-supervised learning via knowledge transfer,'' in {\em Proceedings of the
  IEEE Conference on Computer Vision and Pattern Recognition}, pp.~9359--9367,
  2018.

\bibitem{goyal2019scaling}
P.~Goyal, D.~Mahajan, A.~Gupta, and I.~Misra, ``Scaling and benchmarking
  self-supervised visual representation learning,'' in {\em Proceedings of the
  IEEE/CVF International Conference on Computer Vision}, pp.~6391--6400, 2019.

\bibitem{chen2020big}
T.~Chen, S.~Kornblith, K.~Swersky, M.~Norouzi, and G.~Hinton, ``Big
  self-supervised models are strong semi-supervised learners,'' {\em arXiv
  preprint arXiv:2006.10029}, 2020.

\bibitem{si2020adversarial}
C.~Si, X.~Nie, W.~Wang, L.~Wang, T.~Tan, and J.~Feng, ``Adversarial
  self-supervised learning for semi-supervised 3d action recognition,'' in {\em
  European Conference on Computer Vision}, pp.~35--51, Springer, 2020.

\bibitem{lin2020ms2l}
L.~Lin, S.~Song, W.~Yang, and J.~Liu, ``Ms2l: Multi-task self-supervised
  learning for skeleton based action recognition,'' in {\em Proceedings of the
  28th ACM International Conference on Multimedia}, pp.~2490--2498, 2020.

\bibitem{li2018self}
C.~Li, C.~Deng, N.~Li, W.~Liu, X.~Gao, and D.~Tao, ``Self-supervised
  adversarial hashing networks for cross-modal retrieval,'' in {\em Proceedings
  of the IEEE conference on computer vision and pattern recognition},
  pp.~4242--4251, 2018.

\bibitem{han2020self}
T.~Han, W.~Xie, and A.~Zisserman, ``Self-supervised co-training for video
  representation learning,'' {\em arXiv preprint arXiv:2010.09709}, 2020.

\bibitem{mundhenk2018improvements}
T.~N. Mundhenk, D.~Ho, and B.~Y. Chen, ``Improvements to context based
  self-supervised learning,'' in {\em Proceedings of the IEEE Conference on
  Computer Vision and Pattern Recognition}, pp.~9339--9348, 2018.

\bibitem{wallace2020extending}
B.~Wallace and B.~Hariharan, ``Extending and analyzing self-supervised learning
  across domains,'' in {\em European Conference on Computer Vision},
  pp.~717--734, Springer, 2020.

\bibitem{zhai2019s4l}
X.~Zhai, A.~Oliver, A.~Kolesnikov, and L.~Beyer, ``S4l: Self-supervised
  semi-supervised learning,'' in {\em Proceedings of the IEEE/CVF International
  Conference on Computer Vision}, pp.~1476--1485, 2019.

\bibitem{misra2020self}
I.~Misra and L.~v.~d. Maaten, ``Self-supervised learning of pretext-invariant
  representations,'' in {\em Proceedings of the IEEE/CVF Conference on Computer
  Vision and Pattern Recognition}, pp.~6707--6717, 2020.

\bibitem{bachman2019AmDim}
P.~Bachman, R.~D. Hjelm, and W.~Buchwalter, ``Learning representations by
  maximizing mutual information across views,'' {\em arXiv preprint
  arXiv:1906.00910}, 2019.

\bibitem{chen2021self}
D.~Chen, Y.~Chen, Y.~Li, F.~Mao, Y.~He, and H.~Xue, ``Self-supervised learning
  for few-shot image classification,'' in {\em ICASSP 2021-2021 IEEE
  International Conference on Acoustics, Speech and Signal Processing
  (ICASSP)}, pp.~1745--1749, IEEE, 2021.

\bibitem{he2016deep}
K.~He, X.~Zhang, S.~Ren, and J.~Sun, ``Deep residual learning for image
  recognition,'' in {\em Proceedings of the IEEE conference on computer vision
  and pattern recognition}, pp.~770--778, 2016.

\bibitem{sandler2018mobilenetv2}
M.~Sandler, A.~Howard, M.~Zhu, A.~Zhmoginov, and L.-C. Chen, ``Mobilenetv2:
  Inverted residuals and linear bottlenecks,'' in {\em Proceedings of the IEEE
  conference on computer vision and pattern recognition}, pp.~4510--4520, 2018.

\bibitem{szegedy2016rethinking}
C.~Szegedy, V.~Vanhoucke, S.~Ioffe, J.~Shlens, and Z.~Wojna, ``Rethinking the
  inception architecture for computer vision,'' in {\em Proceedings of the IEEE
  conference on computer vision and pattern recognition}, pp.~2818--2826, 2016.

\bibitem{deng2009imagenet}
J.~Deng, W.~Dong, R.~Socher, L.-J. Li, K.~Li, and L.~Fei-Fei, ``Imagenet: A
  large-scale hierarchical image database,'' in {\em 2009 IEEE conference on
  computer vision and pattern recognition}, pp.~248--255, Ieee, 2009.

\bibitem{DBLP:conf/micad/BarDWG15}
Y.~Bar, I.~Diamant, L.~Wolf, and H.~Greenspan, ``Deep learning with non-medical
  training used for chest pathology identification,'' in {\em Medical Imaging:
  Computer-Aided Diagnosis}, 2015.

\bibitem{lau2018dataset}
J.~J. Lau, S.~Gayen, A.~B. Abacha, and D.~Demner-Fushman, ``A dataset of
  clinically generated visual questions and answers about radiology images,''
  {\em Nature}, 2018.

\end{thebibliography}
\nocite{*}

\end{document}